\documentclass{llncs}
\usepackage{graphicx}
\usepackage{url}
\usepackage{epstopdf}
\usepackage{bm}
\usepackage{multirow}
\usepackage{subfigure}
\usepackage{enumerate}
\usepackage{gensymb}
\usepackage{float}
\usepackage{algpseudocode,algorithm,algorithmicx}
\usepackage{longtable}
\usepackage{amsmath,amssymb,amsfonts}
\usepackage{graphicx}
\usepackage{blindtext}
\usepackage{geometry}
\usepackage{epstopdf} 
 
\begin{document}

\title{Statistical Descriptors-based Automatic Fingerprint
Identification: Machine Learning Approaches}

\author{Hamid Jan\inst{1}, Amjad Ali\inst{2}, Shahid Mahmood\inst{3}*, and Gautam Srivastava\inst{4,5}}

\institute{Department of Electrical Engineering, Sarhad University of Science \& Information Technology, Ring Road,
Peshawar 25000, Kpk, Pakistan; hod.csit@suit.edu.pk \\[1mm]
\and 
Department of Electrical Engineering, Sarhad University of Science \& Information Technology, Ring Road,
Peshawar 25000, Kpk, Pakistan; amjadali@suit.edu.pk\\[1mm]
\and 
Department of Mechanical Engineering, Sarhad University of Science \& Information Technology, Ring Road, Peshawar 25000, Kpk, Pakistan; shahidmahmood757@gmail.com\\[1mm]
\and 
Department of Mathematics and Computer Science, Brandon University, 270 18th Street, Brandon, Canada, R7A 6A9;  srivastavag@brandonu.ca\\[1mm]
\and
Research Center for Interneural Computing, China Medical University, Taichung 40402, Taiwan, Republic of China\\}
\maketitle


\section*{Abstract}
Identification of a person from fingerprints of good quality has been used by commercial applications and law enforcement agencies for many years, however identification of a person from latent fingerprints is very difficult and challenging. A latent fingerprint is a fingerprint left on a surface by deposits of oils and/or perspiration from the finger. It is not usually visible to the naked eye but may be detected with special techniques such as dusting with fine powder and then lifting the pattern of powder with transparent tape. We have evaluated the quality of machine learning techniques that has been implemented in automatic fingerprint identification. In this paper, we use fingerprints of low quality from database DB1 of Fingerprint Verification Competition (FVC 2002) to conduct our experiments. Fingerprints are processed to find its core point using Poincare index and carry out enhancement using Diffusion coherence filter whose performance is known to be good in the high curvature regions of fingerprints. Grey-level Co-Occurrence Matrix (GLCM) based seven statistical descriptors with four different inter pixel distances are then extracted as features and put forward to train and test REPTree, RandomTree, J48, Decision Stump and Random Forest Machine Learning techniques for personal identification. Experiments are conducted on 80 instances and 28 attributes. Our experiments proved that Random Forests and J48 give good results for latent fingerprints as compared to other machine learning techniques and can help improve the identification accuracy.

\section{Introduction}
Over the past century, we have slowly seen the world of fingerprint usage and detection change drastically. From the early days of analog fingerprint usage and storage, to todays world of digital fingerprint usage, storage, and analysis. Fingerprint recognition in today's world relies on a minimum amount of matching between fingerprint data in a fingerprint template and a fingerprint of interest. The template contains a collection of information specifying the type, size, and locations of key features in one or more fingerprints associated with an individual. Fingerprint recognition requires that at least some number of key features in the fingerprint of interest match with the key features stored in the template. However, fingerprints are not always acquired under ideal conditions. Fingerprint images may not contain a sufficient number of key features to allow a good match to a stored fingerprint image~\cite{JainKlareRoss2015}.
Fingerprint recognition can be accomplished using ridge characteristics, minutiae details, using image correlation, or using texture analysis. Minutiae  are  arguably  the  most  important  features  in  fingerprint recognition. In recognition using minutiae, ridge characteristics and the minutiae are required for matching from two images~\cite{MaltoniMaioAk2009}. The information like location, type and direction are obtained. Matching score depends upon the number of corresponding minutiae pairs between the two images which are required to be matched. In the fingerprint recognition using correlation-based matching, the correlation between the corresponding pixels of the two images is computed. This type of matching produces poor results in the presence of noise in the image and the non-linear distortions due to the elastic nature of finger skin. In the texture-based recognition, the information about the pixel level inter-relationship of the image is used to determine the feature set which is then used in the supervised learning of machine based algorithms~\cite{Jiang2000,ShrivastavaSrivastava2014,ZoritaGarciaLlanasRodriguez2001}. 

In recent years, new representations of fingerprint image and new techniques have been proposed to resolve the problems in the fingerprint matching algorithms. In~\cite{GoldRangarajan1996}, the minutiae have been interpreted in the form of graphs. However the high computational complexity of such matching makes its implementation difficult. To overcome this issue, the minutiae are treated as points in the query image and are already stored in the database. The fact that the fingerprint can be seen as a texture oriented system, the texture descriptors can be used to obtain a good representation of the visual content in the image. A global texture descriptor scheme called ``finger code" has been utilized by~\cite{WeiguoGarethMichaelFarzin2007} that employs both the global and local ridge descriptions. A Gabor filter bank is then used for the extraction of these features by measuring the responses of tessellated radial image sectors. In~\cite{JainPrabhakarHongPankanti2000}, the fingerprint texture information is combined with the minutiae details to improve the system performance.
Recently the texture features have been used for fingerprint verification and classification using the statistical descriptors by many researchers \cite{ArivazhaganArulGanesan2007,YazdiGheysari2008,JhatMir2009} whereas the texture is a repeating pattern of local variations in image intensity.
To improve the recognition rate, co-occurrence matrix with distance \cite{JainKlareRoss2015} has been considered and the contrast is calculated from the co-occurrence matrix. The method in this \cite{JhatMir2009} uses the co-occurrence matrix based features for fingerprint classification. The gray level dependence method has been opted by \cite{MohammedKhalilMohamadaKk2010} to extract the texture features of the fingerprint and then the energy has been obtained from co-occurrence matrix as a feature for personal identification.

The remainder of the paper is organized as follows. In Section~\ref{related}, we briefly describe Coherence Diffusion, GLCM, Decision Tree Classifiers. Section~\ref{proposed}, explains the proposed work and the performance evaluators used in this work. Section~\ref{exp} discusses the findings of our experiments. Finally the paper conclusion is given in Section~\ref{conc}.

\section{Related Work}\label{related}
\subsection{Coherence Diffusion}
Coherence-diffusion filtering is the type of filtering which has shown to produce better results for thin and linear structures. Like spatial filtering, it uses neighborhood processing to compute new pixel’s intensities. The filter response varies according to the differential structures within the image. Unlike the linear diffusion filtering in which case the filter produces constant response for the overall image, in the coherence diffusion, the image is smoothed out in the direction of orientation of pixels having similar intensities \cite{AmjadJingZhangNasir2012}. Mathematical description of this process in an image I can be represented using the following equation.

\begin{equation}\label{eq1}
\partial_{t} I=div(D\cdot\nabla I)
\end{equation}
where $\nabla I$ is the image gradient. D and div shows the diffusion tensor and divergence operator respectively.

\subsection{Gray Level Co-Occurrence Matrix (GLCM)}
GLCM has widely been used as a texture feature extraction method since it was first been proposed by Haralick \cite{HaralickShanmuganDinstein1973}. It is a tabulation of how often different combinations of gray levels co-occur in an image or image section. It is used to obtain the relative position of pixels in a neighborhood in an image by estimating the 2nd-order statistical properties of the images.

A co-occurrence matrix $G$ having two pixels intensities m and n with separation distance $d$, co-occurrence direction  and gray level can be expressed mathematically as given below \cite{AmjadXiaojunSaleem2011}.

\begin{equation}
G(m,n,x,y) = \sum_{x=1}^{N} \sum_{y=1}^{N}
\begin{cases}
1, & \text{if }I(x,y)=m\: \text{and}\: I(x+\nabla x,y+\nabla y)=n \\
0, & \text{otherwise}
\end{cases}
\end{equation}
where $\nabla x$ and $\nabla y$ are known as the offsets of the pixels of interest showing the distances between the center pixels and its neighbors.

\subsection{Decision Tree Classifiers}
The decision trees used in machine learning are a guided algorithm for classification. The algorithms of the decision tree used in this work are REPTree, RandomTree, J48, Decision Stump and Random Forest. REPTree which is short for representative tree is a fast-decision tree learner that constructs a decision tree using information gained as a division criterion. The random tree classifier is a collection of tree predictors. It is called random because each tree in the set of trees has the same probability of being sampled. The J48 algorithm considers all possible tests that can divide the data set and selects a test that provides the best information gained. A decision stump is basically a one-level decision tree where the division at the root level is based on a specific attribute / value pair. The decision stumps are single-layer decisions; they are formed quickly compared to the trees. Random Forest is the best to classify large data sets. This classifier can handle thousands of entries without eliminating variables. Random forest generally shows a substantial improvement in yield over the single tree \cite{Zhao2008,Petra2012}.

\section{Proposed Work}\label{proposed}

We propose the following details of the proposed work steps.
\begin{figure}[h!]
	\centering
	\includegraphics[scale=1.3]{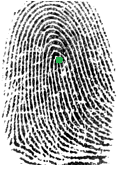}
	\vspace{-.2cm}
	\caption{Fingerprint with Green Spot Showing Core}
	\label{fig01}
\end{figure}

\begin{figure}[h!]
	\centering
	\includegraphics[scale=5.0]{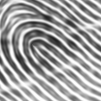}
	\vspace{-.2cm}
	\caption{Fingerprint selected enhanced region}
	\label{fig02}
\end{figure}
\begin{enumerate}

\item The central point of the fingerprint is determined using the Poincare index technique \cite{WangZhangWang2002}. Figure~\ref{fig01} shows a fingerprint image with a central point of green color.

\item The 100x100 dimension region near the central point is found and improved before the feature extraction module, using the diffusion coherence method of image filtering \cite{AmjadJingZhangNasir2012}. Figure~\ref{fig02} shows the selected enhanced region of the fingerprint image obtained after this step.

\item GLCMs having pixel distances of $1$, $2$ and $3$ are obtained in the four directions of $0^\circ$, $45^\circ$, $90^\circ$ and $135^\circ$ for the enhanced image.

\item The following statistical descriptors are calculated from the GLCM to obtain the characteristics vector.
\item \textbf{Variance:} The variance descriptor is the sum of difference between intensity of the central pixel and its 
neighborhood. It reflects the cycle of texture and calculated as

\begin{equation}
\sum_{m} \sum_{n} \bigg[ (m-Avg)^2 \times G(m,n,x,y)\bigg],
\end{equation}

where $m$ and $n$ are the location of pixel, $d$ is the distance between pixels and $\theta$ is angle of pixel orientation. What is AVG????

\item \textbf{Maximum Probability:}
    The maximum probability descriptor is simply the largest entry in the matrix, and corresponds to the strongest response. This could be the maximum in any of the matrices or the maximum overall. This descriptor is calculated as 
\begin{equation}
Max_{m,n} \bigg(\sum_{m} \sum_{n} \bigg[G(m,n,x,y)\bigg]\bigg).
\end{equation}

\item \textbf{Homogeneity:}
    The homogeneity feature of an image is measured by The Inverse Difference moment. This parameter achieves its largest value when most of the occurrences in GLCM are concentrated near the main diagonal. The value of this descriptor is calculated as

\begin{equation}
\sum_{m} \sum_{n}\bigg[\dfrac{G(m,n,x,y)}{1+\lvert m-n \rvert}\bigg].
\end{equation}

\item \textbf{Entropy:}
    The entropy descriptor measures the disorder of an image and it achieves its largest value when all elements in a matrix are equal and is calculated as
\begin{equation}
\sum_{m} \sum_{n}\bigg[G(m,n,x,y) \times log_{10} G(m,n,d,\theta)\bigg].
\end{equation}

\item \textbf{Energy:}
       The energy descriptor which is also called angular second moment and is a measure of textural uniformity, is calculated as 
\begin{equation}
\sum_{m} \sum_{n}\bigg[G^{2}(m,n,x,y)\bigg].
\end{equation}

\item \textbf{Dissimilarity:}
     The contrast descriptor is the difference moment of the matrix and measures the amount of local variations in an image, is calculated as  
\begin{equation}
\sum_{m} \sum_{n}\bigg[(m-n) \times G(m,n,x,y)\bigg].
\end{equation}

\item \textbf{Contrast:}
     The dissimilarity descriptor is similar to contrast but increase linearly. The value is high if the local region has a high contrast. It is calculated as
\begin{equation}
\sum_{m} \sum_{n}\bigg[(m-n)^2 \times  G(m,n,x,y)\bigg].
\end{equation}

\item The set of characteristics is configured in 80 instances and 28 attributes that will be supplied to the machine learning algorithms. 
\item The machine learning techniques, REPTree, RandomTree, J48, Decision Stump and Random Forest are trained and tested with different cross validation folds. 
\item The evaluators for outcome, the instances classification, validity, Recollection, precision and F – measure of the 5 machine learning techniques are found.
\end{enumerate}

\begin{table}[h!]
\centering
\caption{Performance Evaluation of Machine Learning Algorithms}
\begin{tabular}[b]{l@{\hskip 0.35in}l@{\hskip 0.85in}l@{\hskip 0.85in}l@{\hskip 0.55in}}
\hline

\textbf{Classifier} & \textbf{Precision} & 
\textbf{Recall} & \textbf{F-measure} \\\hline
& & & \\
\textbf{J.48} & 0.538 & 0.547 & 0.535   \\\hline
\textbf{Random Forest} & 0.589 & 0.578 & 0.577   \\\hline
\textbf{Random Tree} & 0.39 & 0.406 & 0.396   \\\hline
\textbf{REP Tree} & 0.379 & 0.375 & 0.366   \\\hline
\textbf{Decision Stump} & 0.125 & 0.188 & 0.144   \\\hline
 \\\hline
\end{tabular}
\label{tab01}
\end{table} 

\section{Experimental Results}\label{exp}
The fingerprint images obtained of poor quality from Fingerprint Verification Competition 2002 Database DB1~\cite{MaioMaltoniCappliWaymanJain2002}, have been selected to perform the experiments. DB1 is a collection of 80 fingerprint images from $10$ volunteers each having 8 images. The GLCM is calculated for each of the image enhanced with Coherence filtering method. The seven statistical descriptors are then computed from GLCMs with $1$, $2$ and $3$ inter-pixel distances. These computations of $7$ descriptors in four orientations give rise to a feature set of twenty eight values for fingerprint from an individual. The functions thus obtained are used to train and test the following five machine learning algorithms. The functions are:

\begin{enumerate}[i]
    \item \textbf{REPTree}: stands for Reduces Error Pruning Tree, this algorithm take fast decisions and is found on the concept of calculating the information gain with entropy and minimization of errors as a result of variance.\\

   \item \textbf{Random Tree}: it is a supervised algorithm. It is a collection of learning techniques which further generates learners. It takes features vector as input, classifies it with every tree in forest and finds the class of fingerprint.\\

    \item \textbf{J48}: this algorithm is a simple binary decision tree for classification. The algorithm is tried on each fingerprint in database and the results of its classification are produced.\\

     \item \textbf{Random Forest}: random forest algorithm is best for multidimensional data. It is a mixture of tree predictors and distribution for all trees in a forest. Applying the algorithm on fingerprint, it defines a decision trees and get vote from different decision trees to decide the class of fingerprint. \\

    \item \textbf{Decision Stump}: it is a poor algorithm used for classification. It consists of one level decision tree. Due to its single level decision tree, the performance of Decision Stump is also very bad.
    \end{enumerate}

The techniques are tested using validity, recollection, and F-measure. The instances truly and falsely recognized by the technique are called True Positive and True Negative respectively. 

True Positive describes the values of actual class identified, and the values of identification of predicted class, E.g. if the values indicates that the fingerprint belong to some specific class and the predicted values indicate the same. 

False positive describes the values of actual class not identified and the values of predicted class identified.E.g. If the values indicate that the fingerprint is not of that class but the predicted values indicate that it is of that class.

False Negative describers the values of actual class identified and the predicted class not identified. E.g. If the values indicate that the fingerprint is of that class but the predicted values indicate that it is not of that class. Accuracy, recall and F-measure are described below.

\begin{itemize}
\item Precision= (True Positive)/(True Positive+ False Positive) 					
\item Recall= (True Positive)/(True Positive+False Negative)                                                                                          
\item F-Measure=2*((Precision*Recall) / (Precision + Recall))
\end{itemize}

\begin{figure}[h!]
	\centering
	\includegraphics[scale=0.75]{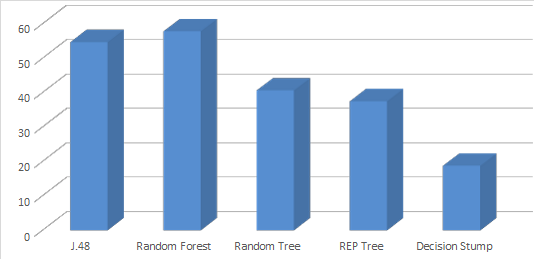}
	\vspace{-.2cm}
	\caption{Performance Measures (Accuracy in $\%$)}
	\label{fig03}
\end{figure}

\begin{figure}[h!]
	\centering
	\includegraphics[scale=0.75]{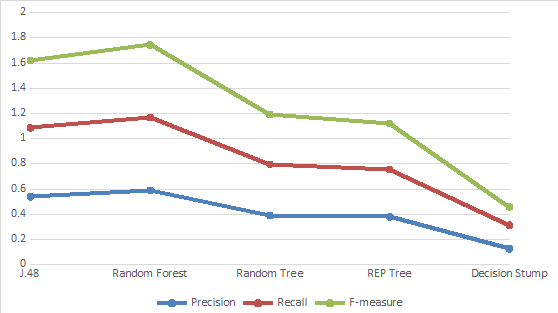}
	\vspace{-.2cm}
	\caption{Performance Measures (Precision, Recall, and F-Measure): Y-axis shows comparative scaling}
	\label{fig04}
\end{figure}

The experimental results shown in Figure~\ref{fig03}, Figure~\ref{fig04} and Table~\ref{tab01} show that \textbf{Random Forest} provides a high percentage of instances correctly classified compared to other proven machine learning algorithms.

The precision, recovery and measurement values of F-Measure obtained for a random forest are the highest among all. The performance of Random Forest is followed by the decision tree classifier J48 with better values than other algorithms. In addition, the experimental results show that the Decision stump of the solution gives incorrect values among the five proven machine learning algorithms.

\section{Conclusion}\label{conc}
In this work, the performance of five state-of-the-art machine learning algorithms in fingerprint recognition for personal identification has been evaluated. Seven significant statistical descriptors are calculated using the GLCM with different pixel spacing distances between pixels in four orientations of public domain DB1 of the FVC 2002 database. These seven descriptors, while calculated in four GLCM orientations, result in a set of characteristics that represent the improved region of the considered fingerprint image. The center point of the fingerprint is identified using the Poincare index method, the high curvature region is selected and enhanced with the diffusion coherence filtering. The experimental results show that the automatic learning algorithms of the random forest and J48 produce better results combined with the random tree, the representation tree and the decision stump. In addition, Random Forest outperforms other proven machine learning algorithms.

\bibliographystyle{splncs03}
\bibliography{ref}

\end{document}